\documentclass[letterpaper]{article}
\usepackage{aaai}
\usepackage{times}
\usepackage{helvet}
\usepackage{courier}
\usepackage{times}
\usepackage{latexsym}
\usepackage{amsmath}
\usepackage{algorithm}
\usepackage{algpseudocode}
\usepackage{url}
\usepackage{multirow}

\usepackage{graphicx}
\usepackage{dblfloatfix} 
\usepackage{caption}
\usepackage[labelformat=simple]{subcaption}

\usepackage{adjustbox}
\usepackage{multirow}

\usepackage{aaai}
\usepackage{times}
\usepackage{helvet}
\usepackage{courier}
\usepackage{times}
\usepackage{latexsym}
\usepackage{amsmath}
\usepackage{algorithm}
\usepackage{algpseudocode}
\usepackage{url}
\usepackage{placeins}
\usepackage[dvipsnames]{xcolor}

\usepackage{multirow}
\usepackage{graphicx}
\usepackage{array}
\usepackage{changepage}
\begin{document}
%
\title{Why Do Masked Neural Language Models Still Need Common Sense Knowledge?}
\author{Sunjae Kwon,\textsuperscript{1}
Cheongwoong Kang,\textsuperscript{2}
Jiyeon Han,\textsuperscript{2}
Jaesik Choi,\textsuperscript{1}\\
\textsuperscript{1}{Graduate School of Artificial Intelligence,  Korea Advanced Institute of Science and Technology}\\
\textsuperscript{2}{School of Electrical \& Computer Engineering, Ulsan National Institute of Science and Technology}\\
sj91kwon@kaist.ac.kr,
cwkang0204@unist.ac.kr, 
jiyeon@unist.ac.kr,
jaesik.choi@kaist.ac.kr}

\maketitle
\begin{abstract}
\begin{quote}
Currently, contextualized word representations are learned by intricate neural network models, such as masked neural language models (MNLMs). The new representations significantly enhanced the performance in automated question answering by reading paragraphs. However, identifying the detailed knowledge trained in the MNLMs is difficult owing to numerous and intermingled parameters. This paper provides empirical but insightful analyses on the pretrained MNLMs with respect to common sense knowledge. First, we propose a test that measures what types of common sense knowledge do pretrained MNLMs understand. From the test, we observed that MNLMs partially understand various types of common sense knowledge but do not accurately understand the semantic meaning of relations. In addition, based on the difficulty of the question answering task problems, we observed that pretrained MLM-based models are still vulnerable to problems that require common sense knowledge. We also experimentally demonstrated that we can elevate existing MNLM-based models by combining knowledge from an external common sense repository.
\end{quote}
\end{abstract}

\section{Introduction}





A long-standing problem and a goal of natural language processing (NLP) is to teach machines to effectively understand language and infer knowledge \cite{winograd1972understanding}. In NLP, reading comprehension (RC) is a task to predict the correct answer in the associated context for a given question. RC is widely regarded as an evaluation benchmark for a machine's ability for the natural language understanding and reasoning \cite{richardson2013mctest}.

Neural language models (NLMs) that consist of neural networks to predict a word sequence distribution have widely been utilized in natural language understanding tasks \cite{radford2018improving}. 
In particular, masked neural language models (MNLMs) including BERT \cite{devlin2019bert}, that are trained to restore randomly masked sequence of words, have recently led to a breakthrough in various RC tasks.
However, the `\textit{black box}' nature of the neural networks prohibits analyzing what type of knowledge leads to performance enhancement and what knowledge remains untrained.


Recently, there has been active research work that tries to explain which type of knowledge is trained on the pretrained NLMs. One common approach is to probe the existence of simple linguistic features such as parts of speech \cite{csahin2019linspector}. 
Previous studies mainly focused on exploring whether the trained model embodies linguistic features for semantic analysis such as tense or named entity recognition (NER), and syntactic analysis such as part-of-speech tagging or dependency parsing for naturally observed texts.
On the contrary, \citet{clark2019does} found that one can well capture syntactic information by observing the self-attention heads' behavior patterns of the BERT.

Common sense knowledge is known to be another essential factor for natural language understanding and reasoning in the RC task \cite{mihaylov2018knowledgeable}. A recent study shows how to attain common sense knowledge from pretrained MNLMs without additional training \cite{feldman2019commonsense}. However, detailed analysis on which knowledge is trained and untrained in the NLMs including MNLMs has not been thoroughly examined to the best of our knowledge.

Our main focus in this paper is to verify how much the MNLM-based RC models answer or process the complicated RC tasks by understanding semantic relations among the words. To address this, we raise the following questions regarding the common sense understanding of MNLMs:

 \begin{enumerate}
    \item Do MNLMs understand various types of common sense knowledge especially relations of attributes? (Section~\ref{sec:knowledge_probing_test})
    \item Do MNLMs understand a relationship between two related relations? (Section~\ref{sec:synonym and antonym})
    \item How do MNLM-based RC models solve problems across different levels of difficulty? (Section~\ref{sec:difficulty_word_overlap})
    \item What are the hardest RC task problems for the MNLM-based RC models? (Section~\ref{sec:question_types})
 \end{enumerate}

For questions 1 and 2, we introduce a new \textit{knowledge probing test} designed to analyze whether the MNLMs understand structured common sense knowledge as semantic triples in an external repository specifically ConceptNet \cite{speer2017conceptnet}. Experimental results on the knowledge probing test reveal that MNLMs partially understand various types of common sense knowledge. However at the same time, we observe unexpected negative results in that MNLMs have a lot of knowledge still untrained yet, and cannot precisely distinguish even the opposite relations.

For questions 3 and 4, we first define the difficulty of an RC problem with the lexical overlapping between the context and the question. Then, we analyze how the MNLMs perform on different levels of difficulty and investigate which type of problems be a critical bottleneck for the current MNLMs. As a result of the analyses, we observe that the lexical variation is a crucial determinant in difficulties of the RC task. In addition, we clarify that the problems that require common sense knowledge are challenging for the MNLM-based RC models.

Based on the above results, we propose a solution that we can ameliorate the limitation of the MNLMs by integrating knowledge originated from an external common sense repository. To verify our solution, we conducted two experiments. 
Firstly, we manually changed the question to integrate the knowledge which is required to solve the problem.
Secondly, we propose a neural network architecture that complements MNLMs with the external common sense repository. In both experiments, we observed that MNLMs could be complemented by integrating common sense knowledge. 

Our main contributions in this paper are as follows:
\begin{itemize}
    \item We proposed a knowledge probing test that measures common sense knowledge trained in MNLM. We observed that MNLMs neither do understand knowledge completely nor precisely. 
    \item By scrutinizing the results of the MNLM based RC models, we observed that current MNLMs have a critical bottleneck with regard to solving problems requiring common sense knowledge. 
    \item We empirically verified that MNLMs can be supplemented by integrating the external common sense repository, manually or automatically.  
\end{itemize}

The paper is organized as follows. Section~\ref{sec:background} briefly describes the notions required to readily understand our paper. Section~\ref{sec:knowledge_probing} introduces our knowledge probing test and demonstrates the results of the test. Then, we present the difficulty levels of the RC problems and the limitations of the MNLM based RC model in Section~\ref{sec:difficulty}. Section~\ref{sec:discussion} discusses what we observed in the previous sections and propose solutions to ameliorate the limitation. Finally, the conclusion is stated in Section~\ref{sec:conclusion}.

\section{Background}
\label{sec:background}

\subsection{Masked Neural Language Models}
In this paper, we address NLMs that calculate a probability distribution over the sequence of words with neural network. Especially, we mainly discuss the BERT that is reffered as the MNLMs in our paper. Two commonly used models of BERT, BERT$_{base}$\footnote{https://storage.googleapis.com/bert\_models/2018\_10\_18/\newline
uncased\_L-12\_H-768\_A-12.zip} and BERT$_{large}$\footnote{https://storage.googleapis.com/bert\_models/2018\_10\_18/\newline
uncased\_L-24\_H-1024\_A-16.zip}, are used in our paper.



\subsubsection{Model Structure}
The BERT model comprises a transformer architecture \cite{vaswani2017attention}. The model has $L$ transformer layers. Each layer comprises $S$ self-attention heads and $H$ hidden dimensions.

\subsubsection{Training Objectives}
BERT is trained to jointly optimize two different losses: 1) masked language model (MLM) loss and 2) next sentence prediction (NSP) loss.

Different from traditional language models that optimize likelihood of the next word prediction, BERT is optimized with the MLM loss. In the MLM loss, tokens in the text are randomly masked with a special token `[MASK]' at a designated proportion, and BERT is optimized with the cross-entropy loss to predict the correct tokens for the masked input.

NSP loss is a binary classification to determine whether a sentence $B$ naturally follows with $A$ in the data sequence. In a positive example, $B$ is sampled in the original context, whereas in a negative example, $B$ is sampled in a randomly selected document. 

\subsubsection{Preprocessing}
For the pretraining, the input of BERT is a conjunction of two sentences
$A$ and $B$. In the sentences, each token is split into a vocabulary, WordPiece \cite{wu2016google}, with 30,000 tokens. In addition, special delimiter tokens `[CLS]' and `[SEP]', that indicate `classification token' and `sentence separate token', respectively, are adopted to integrate two sentences into the following input thread: [CLS]$,A_1,...,A_N,$[SEP]$,B_1...B_M,$[SEP] where $\{A_i\}$ and $\{B_j\}$ are sequential word tokens in sentences $A$ and $B$. 

\subsubsection{Training Data}
BERT is trained on integration of two different corpora: Wikipedia (2,500M words) and Book Corpus (800M words) \cite{zhu2015aligning}. For Wikipedia, only text passages are elicited and hyperlinks are disregarded.

\subsection{Common Sense Repositories}
Before we create common sense queries, determining an external resource where we can extract common senses is necessary. In our paper, we choose ConceptNet, a semantic network widely exploited as a common sense repository in previous studies \cite{weissenborn2017dynamic,wang2018yuanfudao,talmor2019commonsenseqa}.

ConceptNet, part of an open mind common sense (OMCS) \cite{singh2002open} project, is a semantic network designed to help computers understand the words used by people. ConceptNet includes common sense knowledge that originates from several resources: crowdsourcing, expert-creating, and games with a purpose. In our paper, we use ConceptNet 5.6.0 version\footnote{https://s3.amazonaws.com/conceptnet/downloads/2018/edges/\newline
conceptnet-assertions-5.6.0.csv.gz}.

\section{Probing Common Sense Knowledge in MNLMs}
\label{sec:knowledge_probing}
\begin{figure*}[!ht]
  \adjustbox{minipage=2em,raise=-\height}{\subcaption{} \label{fig:L}}%
  \raisebox{-\height}{\includegraphics[width=.95\linewidth]{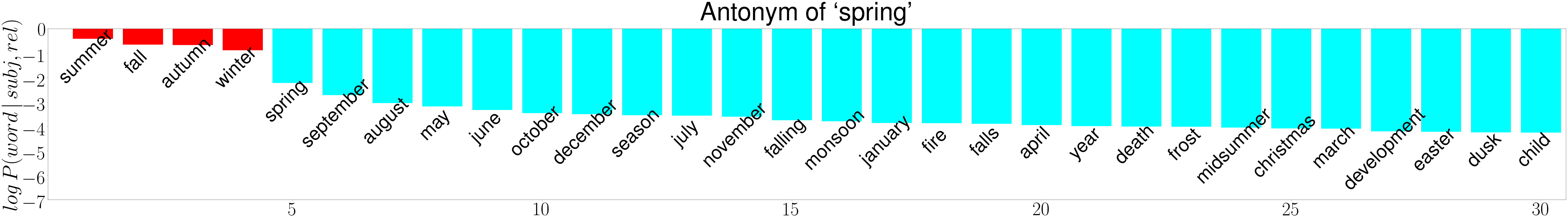}}
  \newline
  \adjustbox{minipage=2em,raise=-\height}{\subcaption{} \label{fig:U}}%
  \raisebox{-\height}{\includegraphics[width=.95\linewidth]{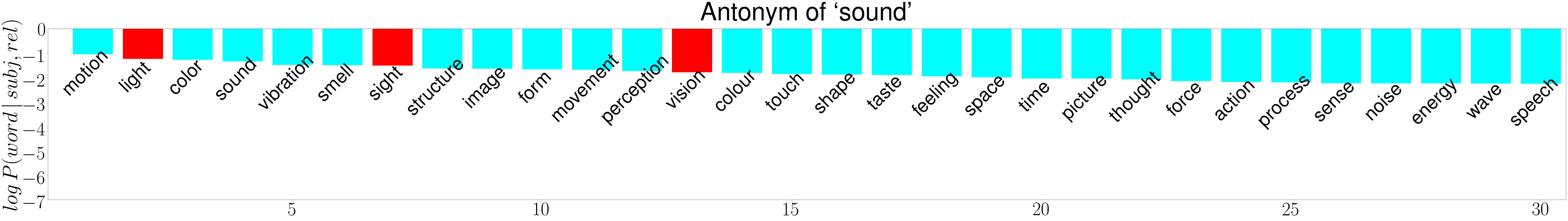}}
  \newline
  \adjustbox{minipage=2em,raise=-\height}{\subcaption{} \label{fig:H}}%
  \raisebox{-\height}{\includegraphics[width=.95\linewidth]{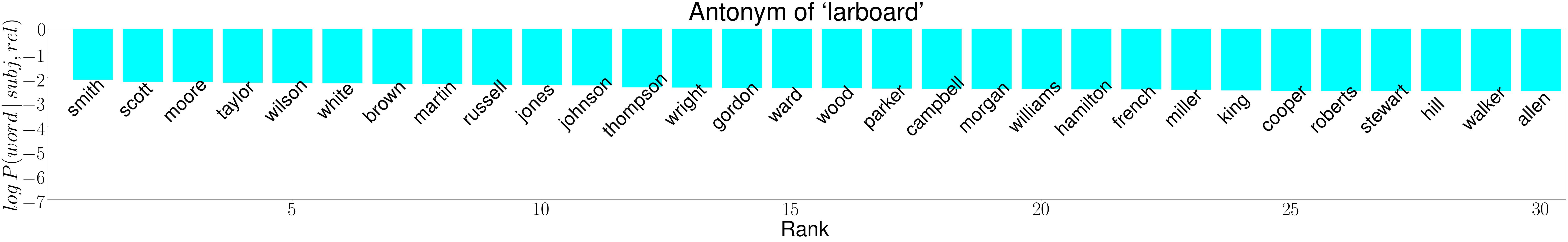}}
  
  \caption{Representative probabilistic distributions of the knowledge probing test results. (a), (b) and (c) show results of `\textit{antonym} of \textit{sound}', `\textit{antonym} of \textit{spring}', and `\textit{antonym} of \textit{larboard}', respectively. The y-axis indicates log$_{10}$ probability and the x-axis denotes the ranking of the words. Correct answers are marked in red.}
  \label{fig:blank_filling_qulitative}
\end{figure*}

This section investigates which types of common sense knowledge are included in the pretrained MNLMs. As the knowledge has a structured form and MNLMs have complex and intermingled attribution, clarifying the trained knowledge is difficult. The Cloze test \cite{chapelle1990cloze}, known to be a reliable assessment for the language ability of a participant, is a task wherein one fills in the correct answer for the blank in the text. In the following example, ``children and \_\_ are opposite .'', the answer word would be `adults' rather than `kids'. To infer the correct answer, we must know not only the meaning of each word but also the semantic relation between the words. Inspired by the Cloze test, we introduce a test called the \textit{knowledge probing test}. 

In the knowledge probing test, we first transform a semantic triple $(s,r,o)$ into a sentence that can be used as an input to a designated MNLM.
We generate sentences through predefined predicate patterns. For example, a predicate pattern of the `Antonym' relation can be ``$s$ and $o$ are opposite .'' 
The predicate patterns of relations are collected from the OMCS dataset \footnote{https://s3.amazonaws.com/conceptnet/downloads/2018/omcs-sentences-more.txt}. All templates used in this paper are provided in Appendix A.

The object in the generated sentence is masked with a special token `[MASK]' such as ``children and [MASK] are opposite.'' MNLMs then try to predict the object from the given subject and relation. We focus on the objects that comprise a single WordPiece token as they are frequently observed in the pretraining corpus. 
\noindent As a result, we can obtain a probabilistic distribution for the masked token and measure the understanding of the MNLMs on common sense knowledge by analyzing the probability of the answer object words.

\subsection{Probing on Various Types of Relations}
\label{sec:knowledge_probing_test}

\begin{figure*}[ht!] 
    \begin{center}
    \includegraphics[width=\linewidth]{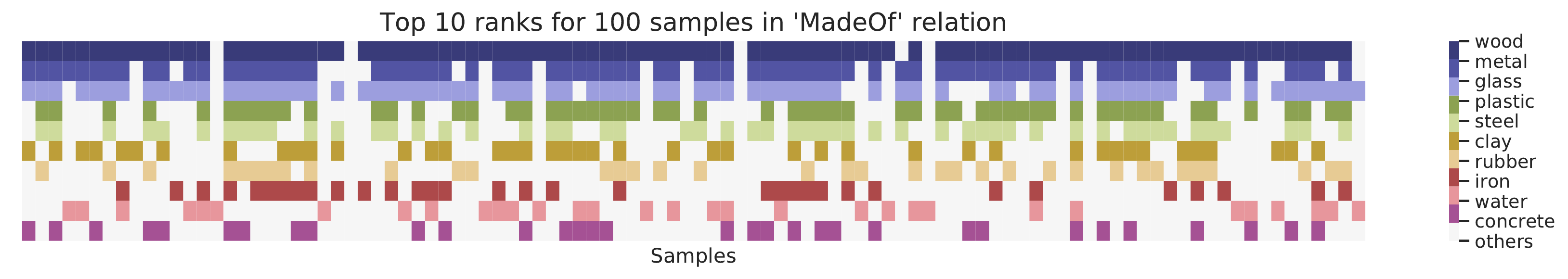}
    \end{center}
\caption{Color coded results on the top 10 words from the model prediction for 100 samples in the 'MadeOf' relation. Colors are labeled with the top 10 most-frequent words. We can notice that top 10 words are redundantly observed in the high rank.}
\label{fig:topk_word_image}
\end{figure*} 
We have conducted the knowledge probing test on 37 relations in ConceptNet to verify whether the MNLMs are properly trained on each relation. The list of 37 relations is provided in Appendix A.

When we visualize the probability distribution on the prediction, we discover that the distributions have roughly three types of aspects. The first type of the distribution shows an \textbf{`L'-shaped} graph, where some words have significantly high probabilities than others. Fig.~\ref{fig:L} is one example of an `L'-shaped distribution. It shows the probability distribution of the predictions for the \textit{`spring's antonym'}. We can see a drastic jump between the probability of `winter' and `spring', which make the figure look similar to the character `L'. 
The second type shows a \textbf{`U'-shaped} graph, where the probabilities smoothly decrease. Fig.~\ref{fig:U} is the distribution for the `sound's antonym', and this is an example of the `U'-shaped graph that shows a smooth curve in the distribution.
The last type shows a \textbf{`--'-shaped} graph, where all candidates share similar probabilities. Fig.~\ref{fig:H} is the distribution of the `larboard's antonym', and the graph looks like a bar. 
We assume the relations that show `L'-shaped graphs are relatively frequently trained on some words as the model is more confident on the words than others. If the model is properly trained, those words with high probabilities will be the answers, as shown in the Fig.~\ref{fig:L}. 
On the contrary, we assume the relations are not trained frequently in the training when the results show `-'-shaped graphs as the model is not as confident on any of its predictions.  

\begin{table}[t!]
\centering
\caption{Results of micro and macro average hits@K for the ConceptNet relations.  
The macro avg. equally average the results of all relations, while the micro avg. weighted average the results of the relations according to the their portion.}
\label{tab:hits@K}
\begin{tabular}{@{}c|c|cccc@{}}
\hline
\multicolumn{2}{c|}{\multirow{2}{*}{Model}} & \multicolumn{4}{c}{Hits@K} \\ \cline{3-6} 
\multicolumn{2}{c|}{} & 1 & 5 & 10 & 100 \\ \hline\hline
\multirow{2}{*}{Micro Avg.} & BERT$_{base}$ & \textbf{5.85} & 10.45 & \textbf{14.32} & \textbf{31.52}\\
 & BERT$_{large}$ & 5.49 & \textbf{10.57} & 14.13 & 30.23  \\\hline
\multirow{2}{*}{Macro Avg.} & BERT$_{base}$ & 5.94 & 13.66 & 17.82 & 38.52\\
 & BERT$_{large}$ & \textbf{7.99} & \textbf{15.00} & \textbf{19.45} & \textbf{41.14} \\ \hline
\end{tabular}
\end{table}

Table~\ref{tab:hits@K} describes the quantitative results of the knowledge probing test measured by the \textit{hits@K} metric \cite{bordes2013translating}. Here, we report micro- and macro-average over the relations. We report individual results on each relation in Appendix B. 
Indeed, large fluctuation can be found in the quantitative results for each relation. Some relations (`DefinedAs', `IsA', ...) show below 20\% in hits@100 while some (`NotCapableOf', `MadeOf', `ReceivesAction') show above 70\%. 
The macro-average displays that BERT$_{large}$ outperforms BERT$_{base}$ while the micro-average shows that BERT$_{base}$ has higher performance than BERT$_{large}$ except for a $hits@5$. This mainly comes from the inconsistency on data distributions between the training dataset of MNLMs and ConceptNet, where some relations occupying a large portion of ConceptNet are not trained more elaborately in the larger model whilst most relations are trained better in the larger model generally. 

Despite the high hit ratios, we suspect that the semantic relations in MNLMs are not as accurate as expected. As an illustrative example, 'MadeOf' relation shows the highest hits@10 with more than 50\% of samples predicting the correct answer within rank 10. However, when we have a closer look at the predictions, some predictions are repeated across the samples. Figure~\ref{fig:topk_word_image} shows the appearance of the 10 frequent words, regardless of order, in the top 10 predictions for 100 samples of the `MadeOf' relation. Regardless of the subject, `wood' appears as a high-rank prediction in most samples. This is followed by `metal' and `glass' as they appear in more than 70\% of samples as high-rank predictions. Herein, we observe that the predictions are biased to the `MadeOf' relation and the conditional effect of the subject is relatively small, leading the model to output the marginal probability of `MadeOf' with general materials. This can be problematic when those frequent words are definitely not the right answer, for example, in cases wherein `wood' is predicted as the most probable answer for the question ``What is butter made of?''. 

\subsection{Probing the Relationship Between Two Relations}\label{sec:synonym and antonym}
In the previous section, we discuss the behavior of MNLMs for each relation but the question ``Do MNLMs precisely understand semantic difference between relations?'' has not been clarified yet. To answer the question, we observe results from the knowledge probing test of opposite relations on the same subject. In particular, we focus on `Antonym' and `Synonym' herein because the two relations are opposite and the sets of correct answers for the two relations are unable to be compatible.
In other words, if the MNLMs precisely understand the meaning of relations, the results of the opposite relation on the same subject should be completely different.

\begin{figure}[!hb]
  \adjustbox{minipage=2em,raise=-\height}{\subcaption{} \label{fig:opposite_test_move}}%
  \raisebox{-\height}{\includegraphics[width=.9\linewidth]{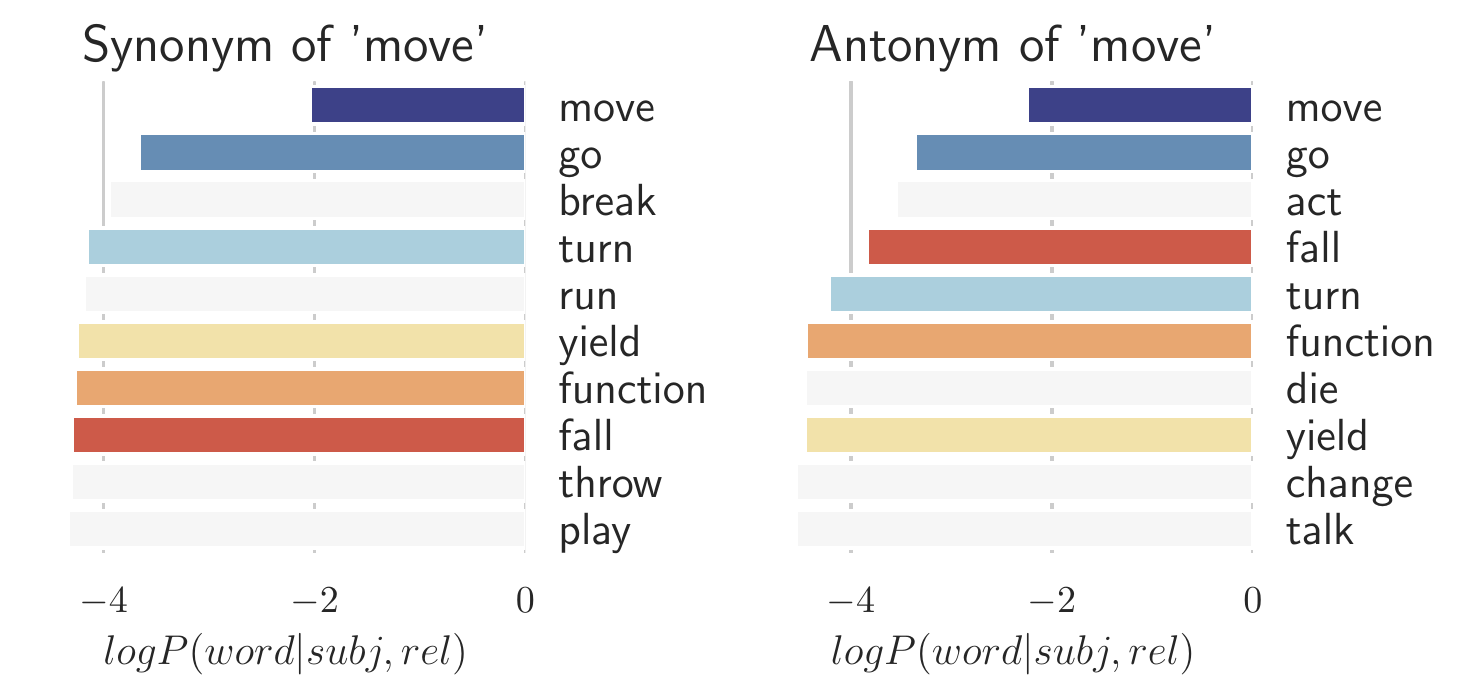}}
  \newline
  \adjustbox{minipage=2em,raise=-\height}{\subcaption{} \label{fig:opposite_test_trust}}%
  \raisebox{-\height}{\includegraphics[width=.9\linewidth]{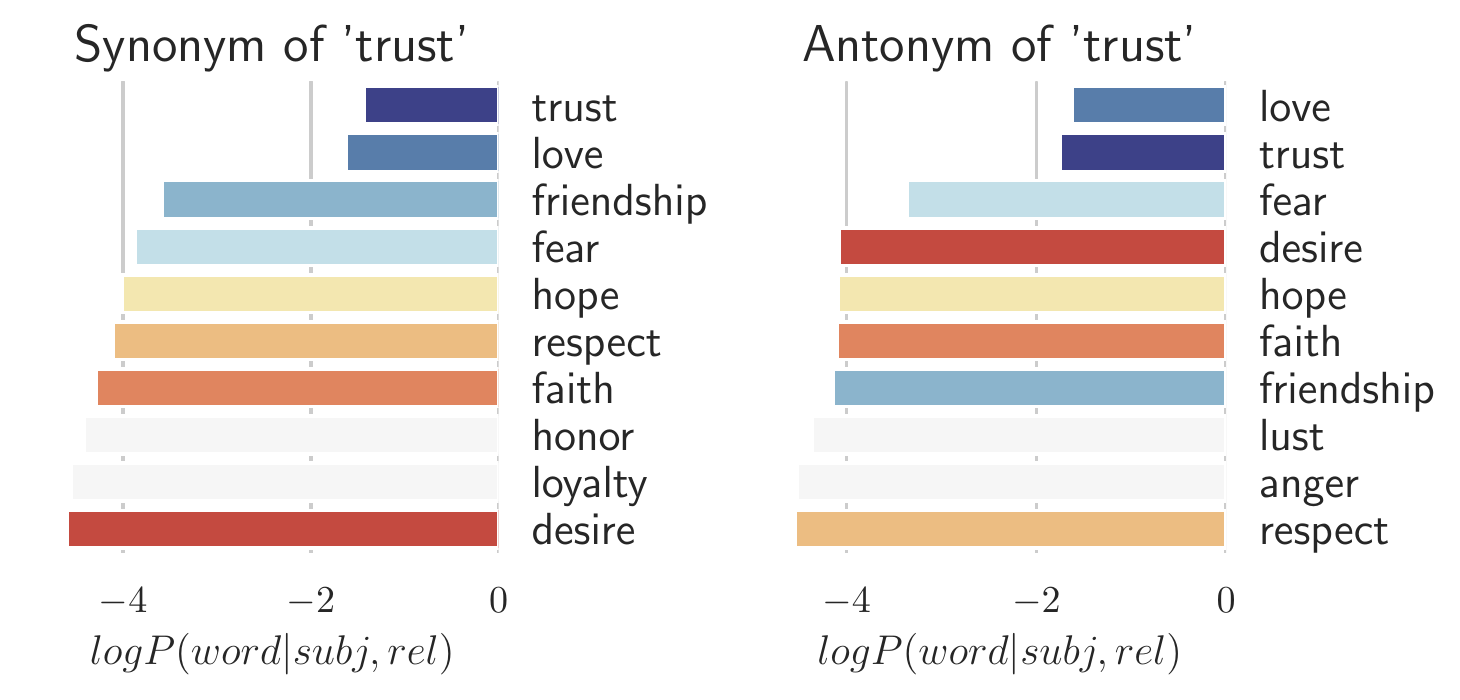}}
  
  \caption{Results on the top 10 words on the opposite relations on subject words a) `move' and b) `trust'. Words commonly observed in both results are painted in the same color, and the other words are in gray.}
  \label{fig:antonym_synonym_overlapping}
\end{figure}
\begin{table}[t]
\centering
\caption{Results of top K overlapping ratio of the `Antonym' and `Synonym' relations.}
\label{tab:overlapping_results}
\begin{tabular}{c|cccc}
\hline
\multirow{2}{*}{Model} & \multicolumn{4}{c}{Overlap@K} \\
\cline{2-5}
 &1 & 5 & 10 & 100 \\
\hline
\hline
BERT$_{base}$ & \textbf{61.19}	&63.47	&64.37	&68.71 \\
BERT$_{large}$ & 58.29	&\textbf{63.60}&	\textbf{64.98}	&\textbf{69.74}\\
\hline
\end{tabular}
\end{table}

\begin{table}[t]
\centering
\caption{Experimental results on the incorrect rate between `Synonym' and `Antonym' relations. }
\label{tab:intergrade_results}
\begin{tabular}{c|c|c|cc}
\hline
\multirow{2}{*}{Model}& \multirow{2}{*}{Template} & \multirow{2}{*}{Answer} & \multicolumn{2}{c}{Hits@K} \\
\cline{4-5}
 & & & 10  & 100 \\
\hline
\hline
\multirow{2}{*}{BERT$_{base}$} & Synonym& Antonym & 31.22 &	55.16 \\
&Antonym & Synonym & 26.25 &	47.18\\\hline
\multirow{2}{*}{BERT$_{large}$} &Synonym & Antonym & 40.18 &	61.92 \\
 & Antonym & Synonym & 25.02 &	48.34 \\
\hline
\end{tabular}
\end{table}
Figure~\ref{fig:antonym_synonym_overlapping} indicates illustrative examples of the opposite relations on the same subject words. Unexpectedly, there are words simultaneously predicted in both `Synonym' and `Antonym'. 
In addition, the quantitative result in Table~\ref{tab:overlapping_results} manifests that the proportion of overlapping words in the opposite relations is rather high ($> 55\%$), making it evident that the MNLMs learn the approximate theme of the opposite relation rather than accurately understand the meaning of the opposite relations.
To demonstrate that high overlapping is undesirable, we measure the incorrect rate by grading the predictions with answers from the opposite relations that are extremely unlikely to be answers. Table~\ref{tab:intergrade_results} lists the measured results. Hits@K in this case can be interpreted as the incorrect rate. As seen from Table~\ref{tab:intergrade_results}, the incorrect rate is rather high in all cases, considering that the no-hit is desirable. Thus, we conclude that MNLMs such as BERT with the current training scheme, do not discriminate opposite relations well.



\section{Analysis on the Reading Comprehension over the Difficulties of the Questions}\label{sec:difficulty}
As reported in the previous section, MNLMs still have incomplete common sense knowledge but MNLM-based RC models outperform existing approaches \cite{radford2018improving,devlin2019bert}.
In this section, we present results on how MNLMs solve RC questions for different difficulty levels (Section~5.1). Subsequently, we report what types of questions are still challenging for the MNLM-based RC models (Section~5.2).

We analyzed BERT$_{base}$, BERT$_{large}$, and a baseline U-Net\footnote{https://github.com/FudanNLP/UNet} model trained on the SQuAD 2.0 RC task dataset \cite{rajpurkar2018know}. This data comprises two types of questions: \textit{has answer} and \textit{no answer}. The \textit{has answer} question contains a contextual answer, whereas the \textit{no answer} question does not have a contextual answer. We train the models with default settings. Finally, as we are unable to access the test set of SQuAD, all analyses are conducted with the development set.

\subsection{ Comparative Study with respect to TF-IDF Similarity} 
\label{sec:difficulty_word_overlap}
\begin{figure}[!ht]
  \adjustbox{minipage=2em,raise=-\height}{\subcaption{} \label{fig:has_answer}}%
  \raisebox{-\height}{\includegraphics[width=.9\linewidth]{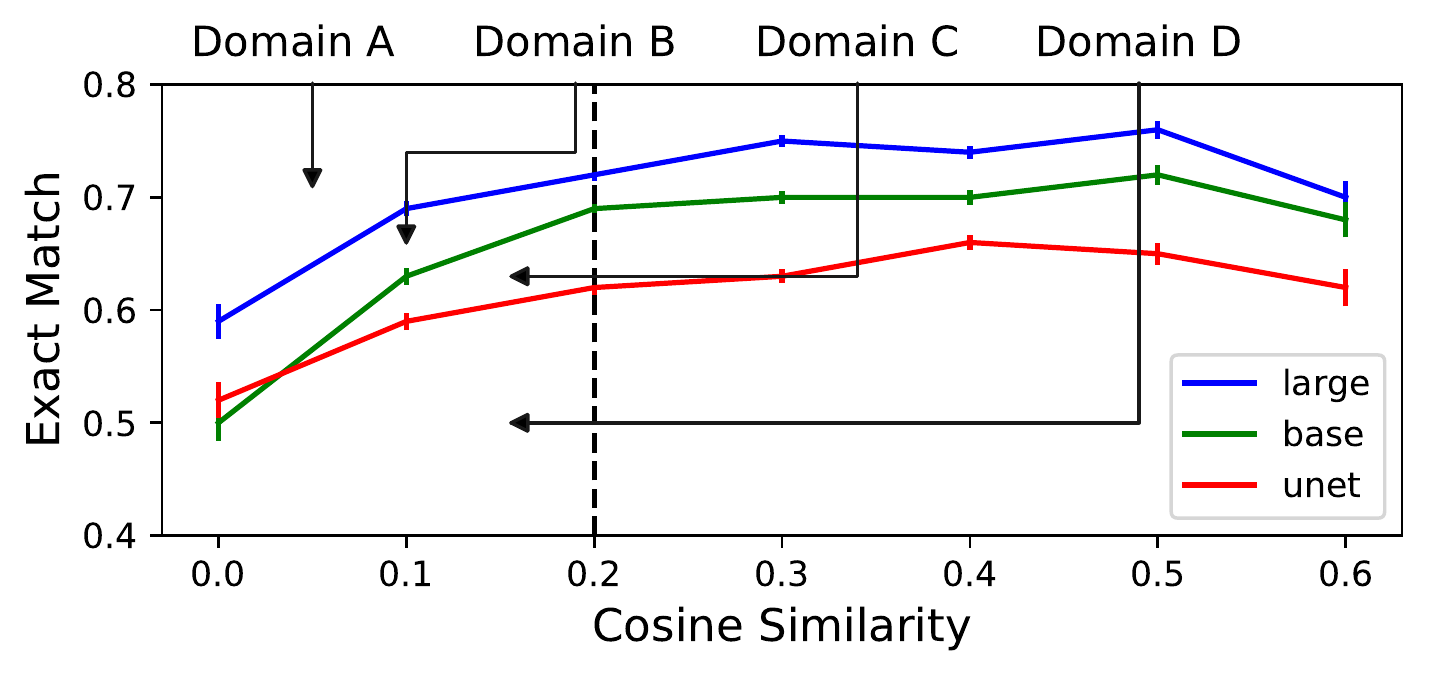}}
  \newline
  \adjustbox{minipage=2em,raise=-\height}{\subcaption{} \label{fig:no_answer}}%
  \raisebox{-\height}{\includegraphics[width=.9\linewidth]{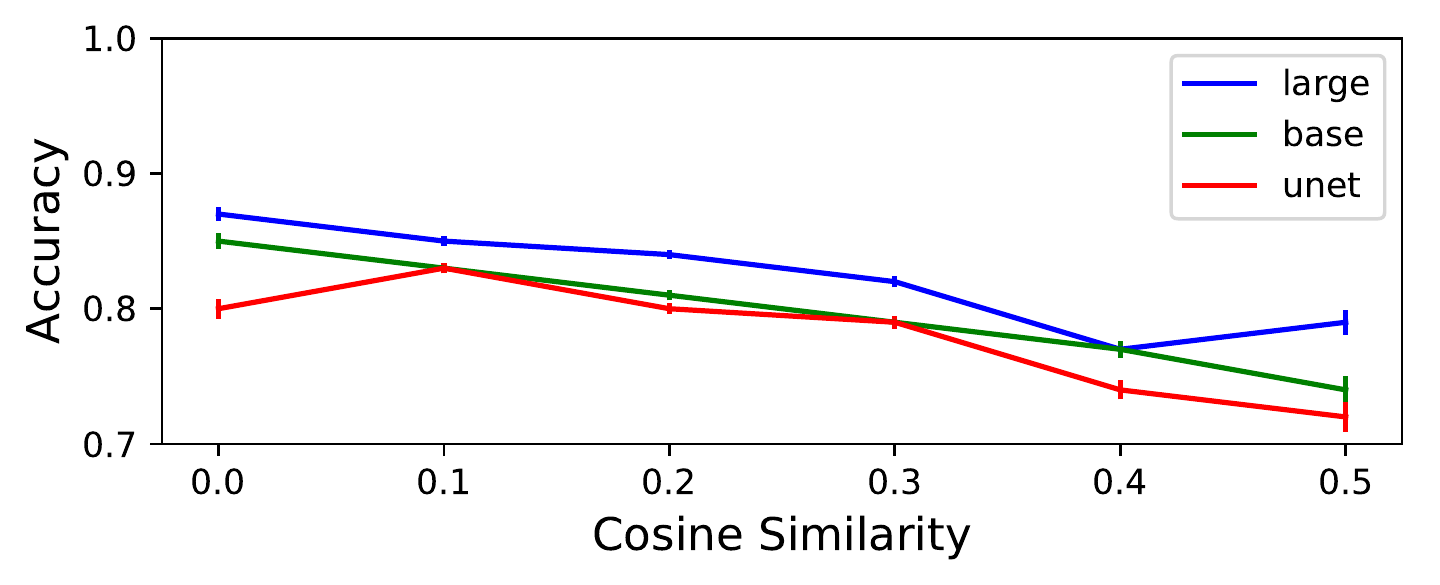}}
  
  \caption{Results on the word overlapping rate and difficulty. X-axis indicates cosine similarity of context and question, and Y-axis denotes its exact matching score. (a) shows results of the \textit{has answer} questions and we marked the domains with the arrow. (b) shows results of the \textit{no answer} questions.}
  \label{fig:word_overlapping_test}
\end{figure}

\begin{table*}[]
\centering
\caption{Question types and their portion on each domain. In the models, Large and Base indicates BERT$_{large}$ and BERT$_{base}$ respectively. There are 6 question categories and the categories can be tagged with duplicates except semantic variation and no semantic variation.}
 \label{tab:question_types}
\begin{tabular}{@{}c|ccc|c|c|c|c|c|c|c@{}}
\hline
\multirow{4}{*}{Domain} & \multicolumn{3}{c|}{\multirow{2}{*}{Models}} & \multicolumn{6}{c|}{Question Type} & \multirow{4}{*}{\begin{tabular}[c]{@{}c@{}}Sampling\\ratio\end{tabular}} \\ \cline{5-10}
 & \multicolumn{3}{c|}{} & \multicolumn{2}{c|}{Semantic Variation} & \multirow{2}{*}{\begin{tabular}[c]{@{}c@{}}Multiple\\Sentence\\Reasoning\end{tabular}} & \multirow{2}{*}{\begin{tabular}[c]{@{}c@{}}No\\Semantic\\Variation\end{tabular}} & \multirow{3}{*}{Others} & \multirow{3}{*}{Typo} &  \\ \cline{2-6}
 & Large & Base & U-Net & Synonymy & \multicolumn{1}{c|}{\begin{tabular}[c]{@{}c@{}}Common Sense \\ Knowledge\end{tabular}} &  &  &  &  &  \\ \hline\hline
A & Fail & Fail & Fail & 33.00\% & \textbf{57.00\%} & \textbf{17.00\%}  & 17.00\% & 2.00\% & \textbf{25.00\%} & 100 / 281 \\
B & Pass & Fail & Fail & \textbf{53.57\%} & 17.86\% & 16.07\% & 32.14\% & 5.36\% & 10.71\%  & 56 / 56 \\
C & Pass & Pass & Fail & 40.45\% & 15.73\% & 16.85\% & 43.82\% & 1.12\% & 6.74\%  & 89 / 89 \\
D & Pass & Pass & Pass & 23.00\% & 12.00\% & 11.00\% & \textbf{65.00\%} & 0.00\% & 3.00\%  &   100 / 531  \\ \hline
\end{tabular}
\end{table*}

We look at the difficulty of the RC problem based on a simple lexical overlapping hypothesis. The hypothesis postulates that the overlap of words in the context and the question strongly correlates with the difficulty level of the RC problem. More specifically, we assume that the \textit{has answer} problem gets easier when the words in the context and question overlap, whereas the \textit{no answer} problem gets harder in the similar situation. 

To verify our assumption, we analyze the relationship between the lexical overlap of context and question, and the performance of RC models. In our case, we calculate the lexical overlap of the context and the question with cosine similarity between TF-IDF term-weighted uni-gram bag-of-words vectors \cite{manning2010introduction}. In addition, we set the performance index of the RC task as an exact matching score and an accuracy value for \textit{has answer} and \textit{no answer} questions. 

Figure~\ref{fig:word_overlapping_test} demonstrates the experimental results for our hypothesis. The results demonstrate that the \textit{has answer} questions tend to be more difficult with less lexical overlapping, whereas the \textit{no answer} shows the opposite tendency. Particularly, there are large performance differences among BERT$_{large}$, BERT$_{base}$, and U-Net in most intervals for the \textit{has answer} questions. In other words, the lexical difference between the question and the context determines the difficulty level of the RC problem.



\subsection{What Types of Questions Are Still Hard for MNLMs?}
\label{sec:question_types}

In this subsection, we analyze which questions account for the performance differences among the RC models. We begin with dividing the \textit{has answer} questions with less lexical overlapping ($similarity < 0.2$), where relatively difficult questions are classified into four domains: A) questions incorrectly answered by all models, B) questions correctly answered only by the BERT$_{large}$, C) questions correctly answered by the BERT$_{base}$ and BERT$_{large}$, and D) questions correctly answered by all models. For each domain, we sample a maximum of 100 questions. Then, by referring the question types in \cite{rajpurkar2016squad}, we categorize each question into the six classes listed in Table~\ref{tab:question_types}. In this case, \textit{synonymy} class means there is a synonym relation between answer sentence and question. The \textit{common sense knowledge} class indicates that common sense is required to solve a question. The \textit{no semantic variation} category denotes that the question has neither synonymy nor common sense knowledge. \textit{Multiple sentence reasoning} class indicates that there are anaphora or clues scattered across multiple sentences. \textit{Others} class indicates that the presented answers have been incorrectly tagged. Finally, the \textit{typo} class denotes a typographical error in a question or answer sentence. Detailed explanations and examples have been provided in Appendix C. 

The experimental results show that the proportion of semantic variation-type questions increases through domain D to A. Especially, the common sense-type questions demonstrate dramatic enhancement in domain A compared to other domains. In addition, the typo-type questions increase significantly in domain A. However, it also manifests that common sense knowledge or typo-type questions are still not handled yet by the MNLM-based RC models.





\section{Discussion and Suggested Solutions}
\label{sec:discussion}
\subsection{What are the Fundamental Limitations of Current MNLM Learning?}
Section~\ref{sec:knowledge_probing} reveals that MNLMs have incomplete information of the common sense knowledge and imprecisely understand the semantic relations, whereas MNLMs can proficiently model contextualized word distribution from the text \cite{tenney2019you}. 

From these findings, we presume that MNLMs learn only 1) the observed information in the corpus and 2) the co-occurrence of the words instead of figuring out semantic relations among the words while they train contextualized word distribution. 
On the contrary, it is difficult for MNLMs to infer the knowledge not observed in the corpus. 

First, MNLMs are incapable of inferring semantic relations that can be inferred from what it already knows. For example, even if an MNLM understands `computer' is made of `transistors' and `transistors' are made of `silicon', it fails to infer `computer' is made of `silicon' if it is hardly found in the corpus.

Second, MNLMs are still naive in understanding the relation over relations. As shown in Section~\ref{sec:synonym and antonym}, the performances of synonym and antonym relations remain almost unchanged even when we grade them interchangeably, after which it is expected to be significantly degraded. This illustrates that MNLMs cannot characterize synonyms and antonyms properly, although they have an obvious relation.


\begin{figure*}[ht!] 
    \begin{center}
    \includegraphics[width=\linewidth]{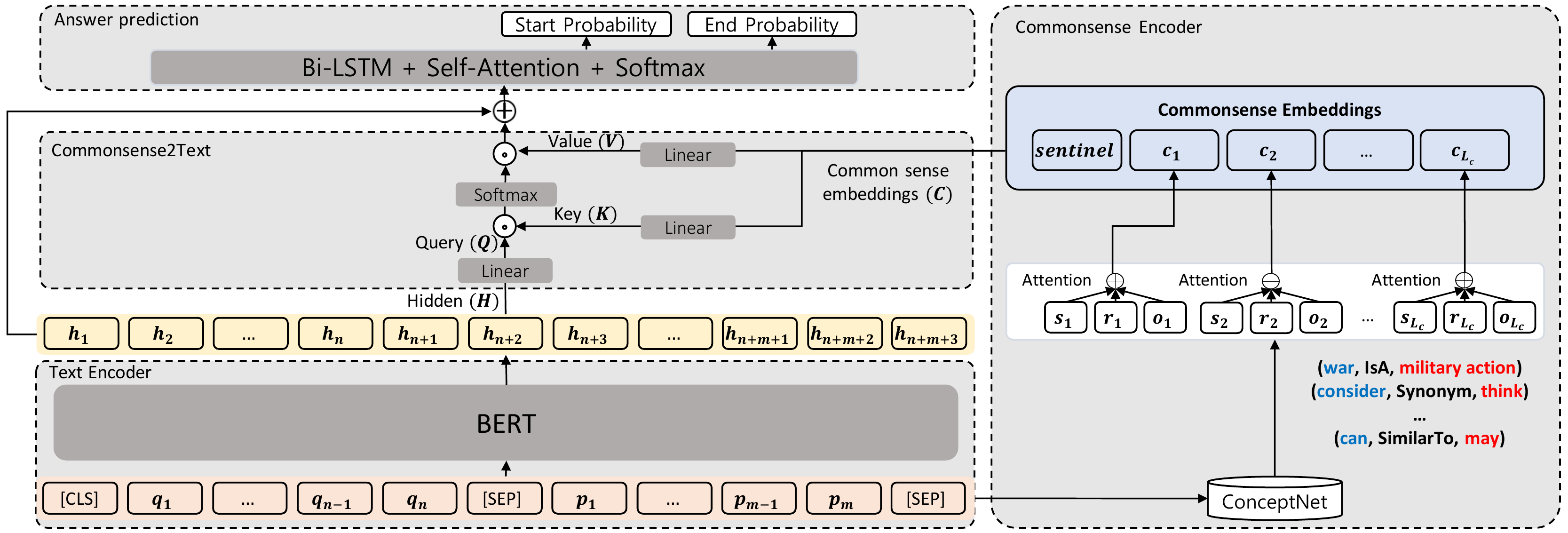}
    \end{center}
\caption{The architecture of our commonsense knowledge incorporated question answering model.}
\label{fig:memory_rc}
\end{figure*}
\subsection{Why do MNLMs Perform Well on QA Tasks without Semantic Relations?}
Despite the limitations, MNLMs such as BERT have shown significant improvements in the RC task. 
However, in Section~\ref{sec:difficulty}, we observed that (1) lexical differences affect the difficulty level of the RC task and (2) there are several RC problems which can be easily solved without extra knowledge. On the contrary, particular problems could not be solved by current MNLM-based RC models and require knowledge on semantic variation. In short, it can be inferred that the semantic variation-type of problems related to synonymy and common sense knowledge remain weak points for the MNLM-based RC models. 

\begin{table}[]
\caption{Empirical analysis on the performances when adapting an external common sense repository. In the table, C2T is an abbreviation of `common sense to text' indicating that we integrate the external common sense repository to the MNLMs.}
\label{tab:memory network}
\begin{tabular}{@{}c|cc|c|cc@{}}
\hline
\multirow{2}{*}{Model} & \multicolumn{2}{c|}{has answer} & no answer & \multicolumn{2}{c}{overall} \\ \cline{2-6} 
  & f1 & exact & accuracy & f1 & exact \\ \hline\hline
BERT$_{base}$  & 73.39 & 67.68 & \textbf{80.10} & 76.75 & 73.90 \\
+ C2T  & \textbf{78.30} & \textbf{72.17} & 77.75 & \textbf{78.02} &\textbf{74.96}\\ \hline
BERT$_{large}$  & 79.33 & 73.48 & 80.89 & 80.11 & 77.28 \\
+ C2T  & \textbf{80.20} & \textbf{74.43} & \textbf{83.53} & \textbf{81.87} &\textbf{78.99}\\ \hline
\end{tabular}
\end{table}

\subsection{MNLMs Need a Learned or Learnable External Common Sense Repository}
\label{sec:complement_NLMs_with_knowledge}

As seen from previous discussions, to implicitly embed common sense knowledge in current MNLM learning, the training data should contain all the common sense knowledge that requires a tremendously large corpus and a large model accordingly. It is almost impossible to gather all common sense knowledge and build such a large model. However, it is obvious that common sense knowledge can help ameliorate the weakness of the current MNLM-based RC models. We try to verify if the external common sense knowledge can be useful for MNLMs in solving RC problems on the hardest problems in domain A. 

\subsubsection{Manually Integrating Common Sense Knowledge}
First, we manually modify the questions to imply knowledge from an external common sense repository and see whether the performance improves. As a result, 56\% of the hardest questions were correctly answered by the BERT$_{large}$ model. 
This shows that we can improve the performance at least for half of those questions from the help of an external common sense repository.  

\subsubsection{Automatically Integrating Common Sense Knowledge}
Secondly, we design a neural memory network that automatically incorporates the repository to the MNLM. 
Figure~\ref{fig:memory_rc} shows the overall model comprising four parts: (1) text encoder, (2) common sense encoder, (3) commonsense2text (C2T), and (4) answer prediction. In the text encoder, the question and context are encoded into the set of hidden vectors $H$ through the BERT. Then, in the common sense encoder, we extract common sense triples that subject and object appear in the text.
Elements of each triple are encoded, then pooled into a single vector through an attention mechanism \cite{bahdanau2014neural}.  
The triple vectors and a sentinel vector, representing the case where there is no relevant knowledge, are gathered to form a common sense embedding $C$.
In the commonsense2text, $C$ is selectively fused into $H$ with the following formula, where $Q$ is a linear transformation of $H$, while $K$ and $V$ are linear transformations of $C$.
\begin{align*}
I = H + Softmax(Q \cdot K) \cdot V  
\end{align*}

\noindent In the answer prediction, a set of knowledge integrated text vectors $I$ is input to the bi-directional long short-term memory (Bi-LSTM) then through self-attention layer and softmax function predicting start and end probabilities of the answer position.   


Table~\ref{tab:memory network} lists experimental results of the MNLMs and our knowledge integrated RC models on SQuAD. The results present that integrating the external common sense repository improves the performance of MNLMs.
We observed that the knowledge integrated BERT$_{large}$ correctly answered 12 out of 100 questions in Domain A of Figure~\ref{fig:has_answer}. Especially, all questions except for one question are common sense knowledge or synonymy types. It implies that coping with external common sense knowledge can be a solution for complementing the weakness of MNLMs. 



\section{Conclusion}
\label{sec:conclusion}
In this paper, we have investigated which types of common sense knowledge are trained in the pretrained MNLMs by proposing a knowledge probing test. We found that MNLMs partially understand some common sense knowledge while the trained knowledge is incomplete and not precise to be distinguished from its opposite. We also analyzed how the MNLM based RC models perform across different difficulty levels of the RC problems and found that questions requiring common sense knowledge are still challenging to current MNLMs. 
Finally, we empirically verified that the limitation of the MNLMs can be overcome by integrating common sense knowledge into the MNLMs.

\section*{Acknowledgement}
This work is supported by IITP grant funded by the Korea government (MSIT) (2017-0-00255, Autonomous digital companion framework and application) and IITP grant funded by the Korea government (MSIT) (2017-0-01779, XAI).

\bibliographystyle{aaai}\bibliography{references.bib}

\onecolumn
\section*{Appendix A: Details on the Templates}
\setcounter{table}{0}
\begin{table*}[!th]
  \centering
  \caption{This table presents details of the templates utilized in our paper. Here, we analyze 37 relations in ConceptNet \cite{speer2017conceptnet}. }
  \label{tab:template}

  \begin{tabular}{c|m{8.5cm}|c}
    \hline
    \textbf{Relation} & \multicolumn{1}{c|}{\textbf{Template}} & \# of samples \\
    \hline\hline
        RelatedTo & [[SUBJ]] is related to [[OBJ]] . & 287,459 \\ 
    HasContext & [[SUBJ]] is used in the context of [[OBJ]] . & 113,066 \\ 
    IsA & [[SUBJ]] is a [[OBJ]] . & 74,316 \\ 
    DerivedFrom & [[OBJ]] is derived from [[SUBJ]] . &  69,510\\     Synonym & [[SUBJ]] and [[OBJ]] are same . & 28,379 \\ 
    FormOf & [[OBJ]] is the root word of [[SUBJ]] . & 27,208\\ 
    EtymologicallyRelatedTo & [[SUBJ]] is etymologically related to [[OBJ]] . & 10,187\\ 
    SimilarTo & [[SUBJ]] is similar to [[OBJ]] . & 8,384 \\ 
    AtLocation & Something you find at [[OBJ]] is [[SUBJ]] . & 7,644 \\ 
    MannerOf & [[SUBJ]] is a way to [[OBJ]] . & 6,230 \\ 
    PartOf & [[SUBJ]] is part of [[OBJ]] . & 5,320 \\ 
    Antonym & [[SUBJ]] and [[OBJ]] are opposite . & 3,932 \\ 
HasProperty & [[SUBJ]] can be [[OBJ]] . & 2,886  \\ 
    UsedFor & [[SUBJ]] may be used for [[OBJ]] . & 2,145 \\ 
    DistinctFrom & [[SUBJ]] is not [[OBJ]] . & 1,256 \\ 
    HasPrerequisite & [[SUBJ]] requires [[OBJ]] . & 1,142 \\ 
    HasSubevent & When [[SUBJ]] , [[OBJ]] . & 1,119 \\ 
    Causes & [[SUBJ]] causes [[OBJ]] . & 999 \\ 
    HasA & [[SUBJ]] contains [[OBJ]] . & 943 \\ 
    InstanceOf & [[SUBJ]] is an instance of [[OBJ]] . & 902 \\ 
    CapableOf & [[SUBJ]] can [[OBJ]] . & 697 \\ 
    ReceivesAction & [[SUBJ]] can be [[OBJ]] . & 658 \\ 
    MotivatedByGoal & You would [[SUBJ]] because [[OBJ]] . & 603 \\ 
    CausesDesire & [[SUBJ]] would make you want to [[OBJ]] . & 556 \\ 
    MadeOf & [[SUBJ]] can be made of [[OBJ]] . & 316 \\ 
    HasLastSubevent & The last thing you do when you [[SUBJ]] is [[OBJ]] . & 302 \\ 
    Entails & [[SUBJ]] entails [[OBJ]] . & 298 \\ 
    HasFirstSubevent & The first thing you do when you [[SUBJ]] is [[OBJ]] . & 280 \\ 
    Desires & [[SUBJ]] wants [[OBJ]] . & 200 \\ 
    NotHasProperty & [[SUBJ]] is not [[OBJ]] . & 161 \\ 
    CreatedBy & [[SUBJ]] is creatd by [[OBJ]] . & 118 \\ 
    DefinedAs & [[SUBJ]] can be defined as [[OBJ]] . & 80 \\ 
    NotDesires & [[SUBJ]] does not want [[OBJ]] . & 71 \\ 
    NotCapableOf & [[SUBJ]] can not [[OBJ]] . & 43 \\ 
    LocatedNear & [[SUBJ]] is typically near [[OBJ]] . & 36 \\ 
    EtymologicallyDerivedFrom & [[SUBJ]] is etymologically derived from [[OBJ]] . & 27 \\ 
    SymbolOf & [[SUBJ]] is an symbol of [[OBJ]] . & 4 \\ 

    \hline
  \end{tabular}
\end{table*}

\newpage
\section*{Appendix B: Qualitative Analysis for Probabilistic Distributions}
\begin{table*}[!ht]
 \centering
  \caption{Results of the $hits@K$ metric for each relation in ConceptNet. }
  \label{tab:results_on_each_relation}
\begin{tabular}{c|cccc|cccc}
\hline
\multicolumn{1}{c|}{\multirow{3}{*}{\textbf{Relations}}} & \multicolumn{8}{c}{$hits@K$} \\ \cline{2-9} 
\multicolumn{1}{c|}{} & \multicolumn{4}{c|}{BERT$_{base}$} & \multicolumn{4}{c}{BERT$_{large}$} \\ \cline{2-9} 
\multicolumn{1}{c|}{} & 1 & 5 & 10 & \multicolumn{1}{c|}{100} & 1 & 5 & 10 & 100 \\ \hline\hline
RelatedTo & 7.60 & 9.30 & 11.77 & 25.38 & 6.51 & 8.50 & 10.97 & 24.14 \\
HasContext & 6.79 & 16.17 & 22.38 & 48.90 & 6.91 & 15.84 & 22.13 & 47.57 \\
IsA & 0.46 & 1.56 & 2.27 & 15.57 & 0.41 & 1.19 & 1.89 & 11.67 \\
DerivedFrom & 0.14 & 5.77 & 10.70 & 31.47 & 0.11 & 3.41 & 6.90 & 23.42 \\
Synonym & 16.16 & 27.33 & 33.12 & 52.70 & 13.38 & 26.74 & 34.69 & 56.39 \\
FormOf & 0.57 & 20.10 & 28.08 & 42.41 & 2.84 & 32.39 & 38.68 & 48.76 \\
EtymologicallyRelatedTo & 5.39 & 8.35 & 10.71 & 22.45 & 3.69 & 6.59 & 9.22 & 21.70 \\
SimilarTo & 1.60 & 4.39 & 6.09 & 14.92 & 2.84 & 7.13 & 10.13 & 23.61 \\
AtLocation & 2.03 & 3.72 & 5.41 & 23.36 & 3.04 & 5.89 & 8.93 & 32.28 \\
MannerOf & 2.66 & 5.05 & 8.77 & 35.71 & 2.17 & 5.85 & 9.61 & 36.25 \\
PartOf & 21.05 & 34.37 & 40.91 & 59.43 & 24.38 & 37.18 & 43.30 & 58.97 \\
Antonym & 17.14 & 25.70 & 32.38 & 53.69 & 28.26 & 34.55 & 40.65 & 63.26 \\
HasProperty & 3.22 & 8.39 & 12.14 & 38.04 & 5.23 & 12.93 & 17.75 & 46.14 \\
UsedFor & 12.87 & 16.50 & 21.44 & 47.16 & 12.26 & 14.78 & 19.25 & 45.72 \\
DistinctFrom & 1.67 & 4.36 & 6.75 & 23.70 & 5.10 & 11.09 & 15.22 & 37.81 \\
HasPrerequisite & 11.30 & 10.56 & 14.73 & 37.29 & 13.75 & 13.35 & 17.93 & 40.54 \\
HasSubevent & 1.79 & 2.55 & 4.03 & 16.20 & 2.32 & 3.39 & 5.11 & 18.40 \\
Causes & 9.71 & 12.73 & 17.05 & 40.79 & 10.81 & 13.90 & 18.65 & 45.81 \\
HasA & 4.24 & 10.55 & 15.17 & 40.35 & 4.67 & 9.75 & 14.19 & 37.22 \\
InstanceOf & 0.00 & 5.93 & 10.29 & 22.43 & 0.11 & 4.92 & 11.12 & 31.92 \\
CapableOf & 10.04 & 17.20 & 24.27 & 53.13 & 12.34 & 22.90 & 28.19 & 52.54 \\
ReceivesAction & 12.01 & 28.12 & 36.51 & 71.44 & 14.89 & 30.52 & 38.85 & 72.45 \\
MotivatedByGoal & 0.00 & 1.07 & 2.37 & 17.90 & 0.00 & 0.17 & 0.76 & 17.74 \\
CausesDesire & 4.32 & 11.52 & 17.59 & 57.25 & 2.34 & 7.54 & 13.95 & 52.13 \\
MadeOf & 12.34 & 44.12 & 51.85 & 72.94 & 18.67 & 42.22 & 50.63 & 75.05 \\
HasLastSubevent & 8.61 & 16.30 & 22.85 & 58.73 & 10.60 & 18.04 & 25.09 & 62.30 \\
Entails & 2.01 & 4.53 & 7.38 & 22.20 & 2.35 & 4.53 & 6.88 & 24.27 \\
HasFirstSubevent & 12.86 & 23.96 & 29.38 & 63.99 & 17.50 & 29.79 & 37.56 & 71.55 \\
Desires & 4.00 & 7.52 & 7.57 & 50.90 & 7.50 & 9.47 & 11.12 & 50.17 \\
NotHasProperty & 4.35 & 14.29 & 18.32 & 42.24 & 6.83 & 23.29 & 27.64 & 60.87 \\
CreatedBy & 2.54 & 9.75 & 15.25 & 35.88 & 0.85 & 5.08 & 10.17 & 29.52 \\
DefinedAs & 0.00 & 2.50 & 3.75 & 17.92 & 2.50 & 4.17 & 10.42 & 33.75 \\
NotDesires & 1.41 & 0.28 & 2.25 & 8.74 & 1.41 & 1.69 & 3.66 & 12.94 \\
NotCapableOf & 16.28 & 32.56 & 41.86 & 73.84 & 18.60 & 27.91 & 40.12 & 76.74 \\
LocatedNear & 2.78 & 8.33 & 13.89 & 36.11 & 5.56 & 8.33 & 8.33 & 25.00 \\
EtymologicallyDerivedFrom & 0.00 & 0.00 & 0.00 & 0.00 & 0.00 & 0.00 & 0.00 & 3.70 \\ 
\multicolumn{1}{c|}{SymbolOf} & \multicolumn{1}{c}{0.00} & \multicolumn{1}{c}{50.00} & \multicolumn{1}{c}{50.00} & \multicolumn{1}{c|}{50.00} & \multicolumn{1}{c}{25.00} & \multicolumn{1}{c}{50.00} & \multicolumn{1}{c}{50.00} & \multicolumn{1}{c}{50.00}\\\hline
\end{tabular}
\end{table*}

\newpage
\section*{Appendix C: Details on the Reading Comprehension Question Types}


\begin{table*}[!ht]
 \centering
  \caption{Examples and descriptions for the question type of the \textit{has answer} questions. The main clues for the categorization of the questions are colored. }
  \label{tab:results_on_each_relation}
\begin{tabular}{c|m{0.30\columnwidth}|m{0.40\columnwidth}}\hline
Question Types &  \multicolumn{1}{c|}{Description} & \multicolumn{1}{c}{Example} \\\hline\hline
Synonymy & There is a clear correspondence between question and context.& \begin{tabular}{@{}p{7cm}@{}}\textbf{Question}: Which entity is the \textbf{\textcolor{red}{secondary}} legislative body?\\\textbf{Context}: ... The \textbf{\textcolor{blue}{second main}} legislative body is the Council, which is composed of different ministers of the member states. ...\end{tabular}  \\\hline
\begin{tabular}[c]{@{}c@{}}Common sense\\ knowledge\end{tabular} & Common sense knowledge is required to solve the question. & \begin{tabular}{@{}p{7cm}@{}}\textbf{Question}: Where is the \textcolor{red}{\textbf{Asian}} influence strongest in Victoria?\\\textbf{Context}: ... Many \textcolor{blue}{\textbf{Chinese}} miners worked in Victoria, and their legacy is particularly strong in Bendigo and its environs. ...\end{tabular}\\\hline
No semantic variation & There is no semantic variation such as synonymy or common sense knowledge. & \begin{tabular}{@{}p{7cm}@{}}\textbf{Question}: Who are the \textbf{\textcolor{red}{un-elected subordinates of member state governments}}?\\\textbf{Context}: ... This means Commissioners are, through the appointment process, the \textbf{\textcolor{blue}{unelected subordinates of member state governments}}. ...\end{tabular} \\\hline
Multi-sentence reasoning & Hints for solving questions are shattered in multiple sentences.  & \begin{tabular}{@{}p{7cm}@{}}\textbf{Question}: Why did \textcolor{red}{\textbf{France}} choose to give up continental lands?\\\textbf{Context}: ...  \textcolor{blue}{\textbf{France}} chose to cede the former, ... \textcolor{blue}{\textbf{They}} viewed the economic value of the Caribbean islands' sugar cane ...\end{tabular} \\\hline
Others & The labeled answer is incorrect. &  \begin{tabular}{@{}p{7cm}@{}}\textbf{Question}: Who \textcolor{red}{\textbf{won the battle}} of Lake George?\\\textbf{Context}: ...  The \textcolor{blue}{\textbf{battle ended inconclusively}}, with both sides withdrawing from the field.  ...\end{tabular} \\\hline
Typo & There exist typing errors in the question or context. & \begin{tabular}{@{}p{7cm}@{}}\textbf{Question}: What kind of measurements define \textbf{\textcolor{red}{accelerlations}}?\\\textbf{Context}... \textbf{\textcolor{blue}{Accelerations}} can be defined through kinematic measurements. ...\end{tabular}\\\hline
\end{tabular}
\end{table*}

\end{document}